\documentclass{article}
\usepackage{ijcai25}

\usepackage{times}
\usepackage{soul}
\usepackage{url}
\usepackage[hidelinks]{hyperref}
\usepackage[utf8]{inputenc}
\usepackage[small]{caption}
\usepackage{graphicx}
\usepackage{amsmath}
\usepackage{amsthm}
\usepackage{booktabs}
\usepackage{algorithm}
\usepackage{algorithmic}
\usepackage[switch]{lineno}
\usepackage{comment}

\usepackage{multicol}
\usepackage{multirow}
\usepackage{makecell}
\usepackage{wrapfig}
\usepackage{tabularx}
\usepackage[capitalize]{cleveref}
\usepackage{xcolor} 
\usepackage{hyperref} 
\usepackage{listings}
\usepackage{enumitem}

\usepackage{mydef}

\hypersetup{
    colorlinks=true, 
    linkcolor=blue,  
    citecolor=blue, 
    urlcolor=blue,  
    pdfborder={0 0 1} 
}

\newcommand{\nibf}[1]{\noindent\textbf{#1}}

\definecolor{color_gan}{RGB}{238, 132, 192}
\definecolor{color_dif}{RGB}{242, 165, 117}
\definecolor{color_oth}{RGB}{232, 229, 240}
\newcommand{\gan}{\textcolor[RGB]{238, 132, 192}}
\newcommand{\dif}{\textcolor[RGB]{242, 165, 117}}

\title{Image Inversion: A Survey from GANs to Diffusion and Beyond}

\author{
Yinan Chen$^{1\star}$\and
Jiangning Zhang$^{1,2\star}$\and
Yali Bi$^3$\and
Xiaobin Hu$^2$\and
Teng Hu$^4$\and\\
Zhucun Xue$^1$\and
Ran Yi$^4$\and
Yong Liu$^{1\dagger}$\And
Ying Tai$^{5}$\\
\affiliations
$^1$College of Control Science and Engineering, Zhejiang University, ~~~ $^2$YouTu Lab, Tencent\\
$^3$College of Computer and Information Science, Southwest University\\
$^4$Department of Computer Science \& Engineering, Shanghai Jiao Tong University\\
$^5$School of Intelligence Science and Technology, Nanjing University\\
\emails
186368@zju.edu.com,
yongliu@iipc.zju.edu.cn,
yingtai@nju.edu.cn
}

\begin{document}

\maketitle

\begin{abstract}
    Image inversion is a fundamental task in generative models, aiming to map images back to their latent representations to enable downstream applications such as editing, restoration, and style transfer. This paper provides a comprehensive review of the latest advancements in image inversion techniques, focusing on two main paradigms: Generative Adversarial Network (GAN) inversion and diffusion model inversion. We categorize these techniques based on their optimization methods. For GAN inversion, we systematically classify existing methods into encoder-based approaches, latent optimization approaches, and hybrid approaches, analyzing their theoretical foundations, technical innovations, and practical trade-offs. For diffusion model inversion, we explore training-free strategies, fine-tuning methods, and the design of additional trainable modules, highlighting their unique advantages and limitations. Additionally, we discuss several popular downstream applications and emerging applications beyond image tasks, identifying current challenges and future research directions. By synthesizing the latest developments, this paper aims to provide researchers and practitioners with a valuable reference resource, promoting further advancements in the field of image inversion. 
    We keep track of the latest works at \href{https://github.com/RyanChenYN/ImageInversion}{\textcolor{magenta}{https://github.com/RyanChenYN/ImageInversion}}.
\end{abstract}

\section{Introduction} \label{sec:introduction}

\begin{figure*}[thp]
    \centering
    \includegraphics[width=1\linewidth]{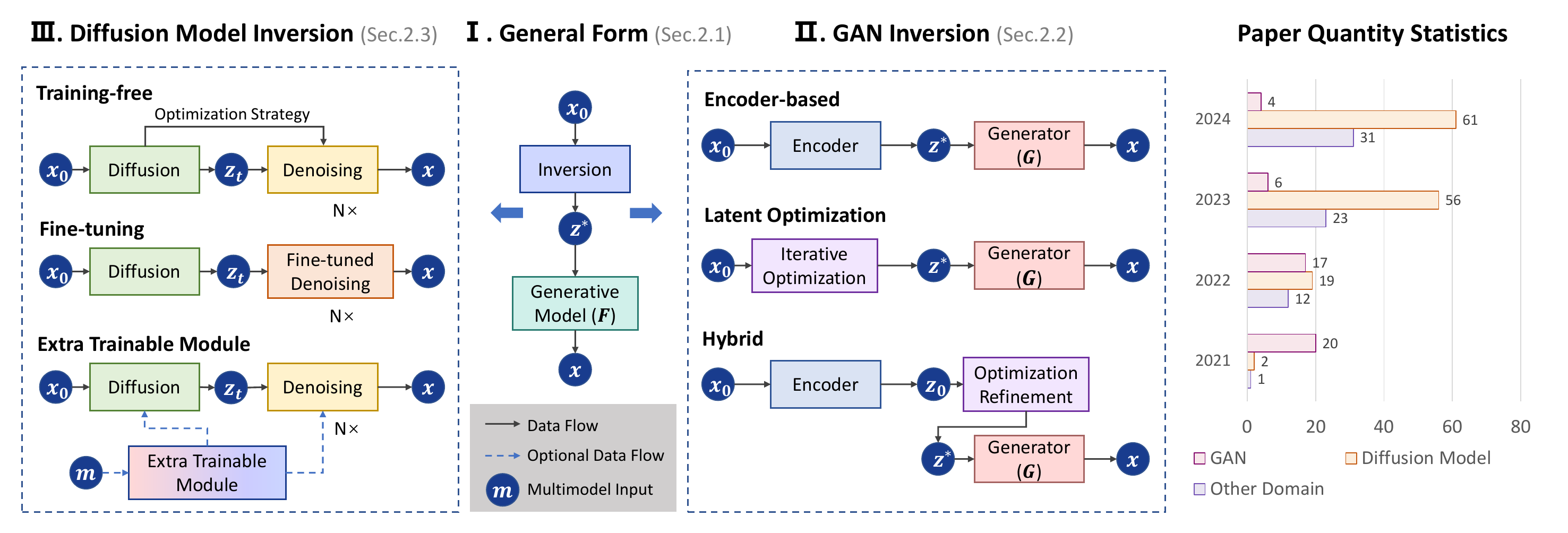}
    \vspace{-2.0em}
    \caption{
    \textbf{Left:} Diagrammatic overview of the general formulaic description of image inversion (\textbf{\uppercase\expandafter{\romannumeral1}}), as well as its instantiation in GAN (\textbf{\uppercase\expandafter{\romannumeral2}}) and Diffusion (\textbf{\uppercase\expandafter{\romannumeral3}}) frameworks. 
    \textbf{Right:} Works summarization on different frameworks in recent years. 
    Only the works from the past four years are listed. Due to the superior performance of diffusion models, the interest in GAN-based work has been declining annually. 
    }
    \vspace{-1.5em}
    \label{fig:teaser}
\end{figure*}

Image inversion refers to the task of mapping a given image back to the latent representation with a pre-trained generative model. This task holds significant importance in applications such as image editing, style transfer, image restoration, \textit{etc}~\cite{survey1,survey2}. Through inversion techniques, users can effectively leverage the rich semantic information of generative models to achieve efficient control and modification of real images, making it an increasingly independent and active research direction.

Early researches on image inversion began with the rise of GANs~\cite{iGAN}, focusing primarily on how to project images into the latent space of GANs to facilitate subsequent image editing and generation tasks. The advent of the StyleGAN series~\cite{StyleGAN,StyleGAN2} notably improved the accuracy and efficiency of image inversion techniques. However, these methods have certain limitations~\cite{e4e,PTI,SDIC}: encoder-based one-forward approaches still yield suboptimal results, while optimization-based methods are time-consuming and fail to meet the demands of general image editing and high-precision applications, such as portrait photography. 
In recent years, diffusion models have emerged as a new favorite in the field of generative models due to their powerful generative capabilities and stable training processes. From DDPM~\cite{DDPM}/DDIM~\cite{DDIM} to LDM~\cite{LDM}, open-source models like the Stable Diffusion series have significantly enhanced the controllability and effectiveness of image editing, leading to numerous excellent training-free and fine-tuning solutions~\cite{Negative-prompt,StyleID,FreeControl}. Recent breakthroughs, such as the DiT~\cite{DiT} framework and flow matching techniques, have provided new insights and methods for image inversion. The diverse development from GANs to diffusion models has also laid the foundation for high-fidelity image inversion tasks and controllable editing applications in complex scenarios.

This survey systematically reviews and summarizes the development trajectory of these technologies, abstractly defining the problem from a formulaic perspective, and delving into the principles and practical issues of different categories of methods. It comprehensively covers image inversion and related subfields, providing a thorough discussion.

\nibf{Scope.} 
This paper focuses on the two main frameworks for image inversion: GAN inversion and diffusion model inversion. For GAN inversion, we conduct a comprehensive analysis and comparison from three perspectives: Encoder-based Approaches, Latent Optimization Approaches, and Hybrid Approaches. For diffusion model inversion, we categorize the methods from a training perspective into Training-free Approaches, Fine-tuning Approaches, and Extra Trainable Module Approaches, discussing the advantages and disadvantages of each. Additionally, we analyze the latest technological trends, such as DiT-based inversion methods\cite{DiT4Edit}, and explore the applications of inversion techniques in images and broader fields like video\cite{Videoshop} and audio\cite{ZETA}. This survey primarily analyzes work post-2021 to ensure its relevance and forward-looking nature. Due to space constraints, only representative works are discussed, with comprehensive and up-to-date work continuously tracked on this \href{https://github.com/RyanChenYN/ImageInversion}{\textcolor{magenta}{project page}}.

\nibf{Discussion with Related Surveys.}
Compared to existing surveys, such as ~\cite{survey1} which focuses on early GAN-based approaches, and recent work ~\cite{survey2} which focuses on diffusion-based approaches, this survey integrates GAN inversion and diffusion model inversion into a unified framework for systematic comparison, filling a research gap in this field. It also extends the discussion of inversion to non-image applications, providing readers with a more comprehensive perspective.

\nibf{Contributions.}
Firstly, we provide a thorough review of the latest advancements in the field of image inversion, covering the key inversion techniques of the two main generative models (GANs and diffusion models). By systematically categorizing these methods, we reveal the intrinsic connections and technical differences, offering clear theoretical guidance for researchers. Secondly, we discuss the primary applications from an image-level perspective and related field advancements. Finally, we summarize the current challenges in the research and propose a series of potential future research directions, providing important references for further development in the field of image inversion.

\section{Taxonomy of Image Inversion Approaches} \label{sec:method}

\tikzstyle{leaf}=[draw=black,
    rounded corners,minimum height=1em,
    text width=24.50em, edge=black!10, 
    text opacity=1, align=center,
    fill opacity=.3,  text=black,font=\scriptsize,
    inner xsep=3pt, inner ysep=1pt,
    ]
\tikzstyle{leaf1}=[draw=black,
    rounded corners,minimum height=1em,
    text width=6.28em, edge=black!10, 
    text opacity=1, align=center,
    fill opacity=.5,  text=black,font=\scriptsize,
    inner xsep=3pt, inner ysep=1pt,
    ]
\tikzstyle{leaf2}=[draw=black, 
    rounded corners,minimum height=1em,
    text width=6.28em, edge=black!10, 
    text opacity=1, align=center,
    fill opacity=.8,  text=black,font=\scriptsize,
    inner xsep=3pt, inner ysep=1pt,
    ]

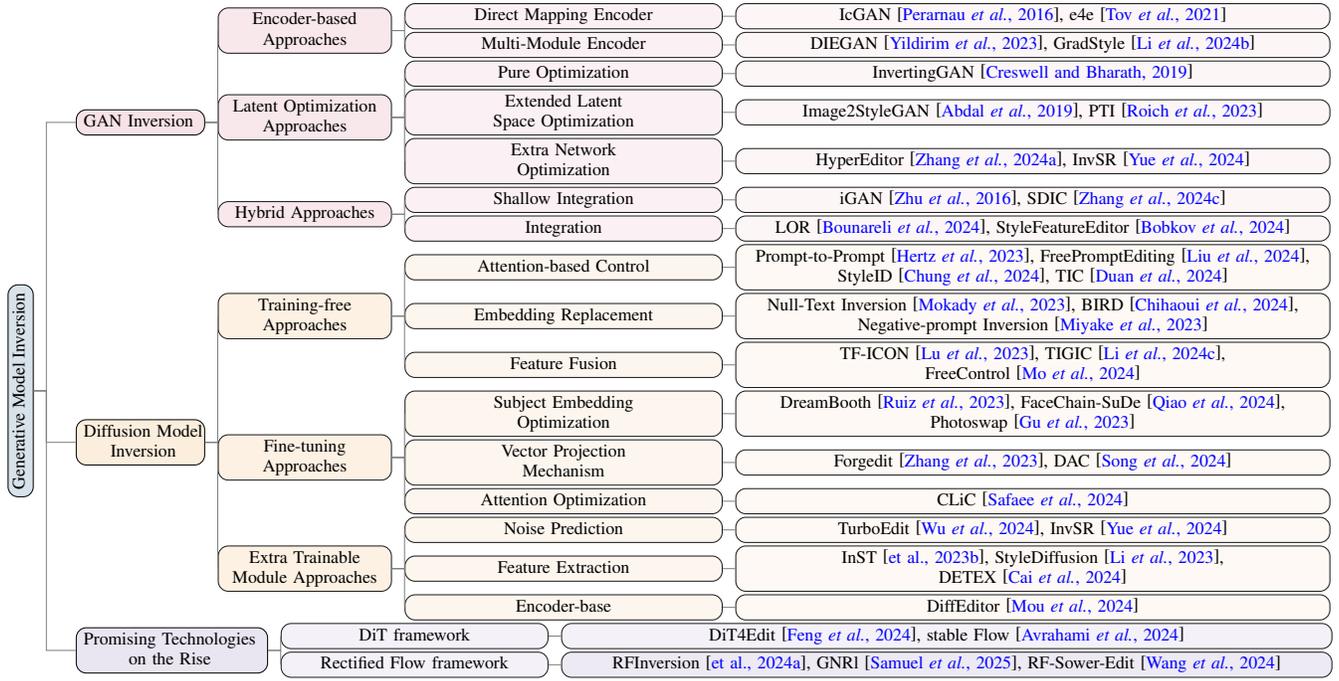
\begin{figure*}[ht]
\centering
\resizebox{1.00\linewidth}{!}{
\begin{forest}
for tree={
    forked edges,
    grow=east,
    reversed=true,
    anchor=base west,
    parent anchor=east,
    child anchor=west,
    base=middle,
    font=\scriptsize,
    rectangle,
    rounded corners=0.1pt, 
    draw=black, 
    edge=black!50, 
    rounded corners,
    align=center,
    minimum width=2em, 
    l sep=5pt,
    s sep=1pt,
    inner xsep=3pt, 
    inner ysep=1pt,
  },
  where level=1{text width=4em}{}, 
  where level=2{text width=1em,font=\scriptsize}{},
  where level=3{text width=8em,font=\scriptsize}{},
  where level=4{text width=10em,font=\scriptsize}{},
  where level=5{text width=1em,font=\scriptsize}{},
  [Generative Model Inversion,rotate=90,anchor=north,edge=black!50,fill=myblue,draw=black
    [GAN Inversion,edge=black!50,text width=4.5 em, fill=myred
        [Encoder-based \\ Approaches, leaf2, fill=myred
            [Direct Mapping Encoder, leaf1,text width=12 em,fill=myred
                 [{IcGAN~\cite{IcGAN}, e4e~\cite{e4e}},leaf,text width=23 em, fill=myred]
            ]
            [Multi-Module Encoder, leaf1,text width=12 em,fill=myred
                 [{DIEGAN~\cite{DIEGAN}, GradStyle~\cite{GradStyle}},leaf,text width=23 em,fill=myred]
            ]
        ]
        [Latent Optimization \\ Approaches, leaf2,fill=myred
            [Pure Optimization, leaf1,text width=12 em, fill=myred[InvertingGAN~\cite{InvertingGAN}, leaf,text width=23 em, edge=black!50, fill=myred]
            ] 
            [Extended Latent \\Space Optimization, leaf1,text width=12 em,fill=myred
                 [{Image2StyleGAN~\cite{Image2StyleGAN}, PTI~\cite{PTI}} ,
leaf,text width=23 em,fill=myred]
            ] 
            [Extra Network \\Optimization, leaf1, text width=12 em, fill=myred
                 [{HyperEditor~\cite{HyperEditor}, InvSR~\cite{InvSR}}, leaf,text width=23 em,fill=myred]
            ]  
        ]
        [Hybrid Approaches, leaf2,fill=myred
            [Shallow Integration, leaf1,text width=12 em, fill=myred
                 [{iGAN~\cite{iGAN}, SDIC~\cite{SDIC}},leaf,text width=23 em,fill=myred]
            ]  
            [Integration, leaf1,text width=12 em, fill=myred
                 [{LOR~\cite{LOR}, StyleFeatureEditor~\cite{StyleFeatureEditor}},leaf,text width=23 em,fill=myred]
            ]  
        ]
    ]
    [Diffusion Model \\ Inversion,edge=black!50,text width=4.5em, fill=myyellow
    	[Training-free \\ Approaches,leaf2, fill=myyellow,
            [Attention-based Control, leaf1, text width=12 em, fill=myyellow
                 [{Prompt-to-Prompt~\cite{p2p}, FreePromptEditing~\cite{FreePromptEditing}, \\
                 StyleID~\cite{StyleID}, TIC~\cite{TIC}},leaf,text width=23 em,fill=myyellow]
            ]  
            [Embedding Replacement, leaf1,text width=12 em, fill=myyellow
                 [{Null-Text Inversion~\cite{Null-Text}, BIRD~\cite{BIRD}, \\ Negative-prompt Inversion~\cite{Negative-prompt}},leaf,text width=23 em,fill=myyellow]
            ]  
            [Feature Fusion, leaf1,text width=12 em, fill=myyellow 
                 [{TF-ICON~\cite{TF-ICON}, TIGIC~\cite{TIGIC}, \\ FreeControl~\cite{FreeControl}},leaf,text width=23 em,fill=myyellow]
            ]  
         ]
         [Fine-tuning \\Approaches, leaf2, fill=myyellow
            [Subject Embedding \\Optimization, leaf1,text width=12 em,fill=myyellow
                 [{DreamBooth~\cite{DreamBooth}, FaceChain-SuDe~\cite{FaceChain-SuDe},\\ Photoswap~\cite{Photoswap}}, leaf,text width=23 em,fill=myyellow]
            ]
            [Vector Projection \\ Mechanism, leaf1, text width=12 em, fill=myyellow
                 [{Forgedit~\cite{Forgedit}, DAC~\cite{DAC}},leaf,text width=23 em,fill=myyellow]
            ]
            [Attention Optimization, leaf1, text width=12 em, fill=myyellow
                 [CLiC~\cite{CLiC},leaf,text width=23 em,fill=myyellow]
            ]
         ]
         [Extra Trainable\\ Module Approaches, leaf2, fill=myyellow
            [Noise Prediction, leaf1,text width=12 em, fill=myyellow
                 [{TurboEdit~\cite{TurboEdit}, InvSR~\cite{InvSR}},leaf,text width=23 em,fill=myyellow]
            ]
            [Feature Extraction , leaf1,text width=12 em, fill=myyellow
                 [{InST~\cite{InST}, StyleDiffusion~\cite{StyleDiffusion},\\DETEX~\cite{DETEX}},leaf,text width=23 em,fill=myyellow]
            ]
            [Encoder-base, leaf1, text width=12 em, fill=myyellow
                 [DiffEditor~\cite{DiffEditor} ,leaf,text width=23 em,fill=myyellow]
            ]
         ]
    ]
    [Promising Technologies\\ on the Rise, edge=black!50,text width=7em, fill=mypurple
    [DiT framework,leaf1, text width=10 em, fill=mypurple
    [{DiT4Edit~\cite{DiT4Edit}, stable Flow~\cite{Stable}},leaf1,text width=30 em, fill=mypurple]
    ]
    [Rectified Flow framework,leaf1, text width=10 em,fill=mypurple
    [{RFInversion~\cite{RFInversion}, GNRl~\cite{GNRI}, RF-Sower-Edit~\cite{RF-Solver-Edit}},leaf2,text width=30 em, fill=mypurple]]
    ]
  ]
\end{forest}
}
\caption{A taxonomy of generative model inversion approaches from GANs to Diffusion and beyond.}
\label{fig:taxonomy_of_GMI}
\vspace{-1.5em}
\end{figure*}

\subsection{Background and Definition of Image Inversion}
\nibf{Generative Model Inversion} 
refers to the process of recovering the latent variables of a pretrained generative model given an observed output \(x_0\). As shown in ~\cref{fig:teaser}-\textbf{\uppercase\expandafter{\romannumeral1}}, suppose we have a generative model $F$ that maps a latent code \(z\) from a latent space \(Z\) to the output image space \(X\): 
\begin{equation}
F: Z \to X, \quad x = F(z).
\end{equation} 
Given an input image \(x_0 \in X\), the goal of inversion is to find a latent code \(z^* \in Z\) such that the reconstructed image \(x = F(z^*)\) is as close as possible to the input \(x_0\). This process can be formulated as the following optimization problem:

\begin{equation}
z^* = \underset{z}{\mathrm{argmin}} \, \mathcal{L}(F(z), x_0),
\end{equation} 
where \( \mathcal{L} \) is a predefined loss function that measures the discrepancy between the reconstructed image \(x = F(z)\) and the original input \(x_0\). Common choices for \( \mathcal{L} \) include pixel-wise, perceptual, and SSIM losses.

\nibf{GAN Inversion} 
is a specific instance of generative model inversion. As show in ~\cref{fig:teaser}-\textbf{\uppercase\expandafter{\romannumeral2}}, the generator is denoted as $G$, which maps latent variables \(z\) from the latent space \(Z\) to the output space \(X\): $G: Z \to X$, where $x = G(z).$

Its goal is to find a latent code \(z^* \in Z\) such that the reconstructed image \(x = G(z^*)\) is as close as possible to the input \(x_0\), expressed as the following optimization problem:
\vspace{-0.2em}
\begin{equation}
z^* = \underset{z}{\mathrm{argmin}} \, \mathcal{L}(G(z), x_0),
\end{equation} 
\vspace{-0.2em}
where \( \mathcal{L} \) measures the discrepancy between the reconstructed image \(x = G(z)\) and the input \(x_0\). In GAN inversion, additional regularization terms may be introduced to ensure the latent code \(z^*\) remains semantically meaningful or lies within the generator's latent space distribution.

\nibf{Diffusion Model Inversion}
models are stepwise generative models that create data \(x_0\) by iteratively denoising an initial noise \(z_T\). As shown in ~\cref{fig:teaser}-\textbf{\uppercase\expandafter{\romannumeral3}}, the generative process of a diffusion model is described as:  
\vspace{-0.5em}
\begin{equation}
\quad p_\theta(x_{0:T}) = p(z_T) \prod_{t=1}^T p_\theta(z_{t-1} | z_t),
\end{equation} 
where \(z_t\) represents the intermediate latent variable at timestep \(t\), \(z_T\) is the initial noise, \(T\) is the total number of timesteps, and \(p_\theta(z_{t-1} | z_t)\) denotes the conditional denoising probability parameterized by \(\theta\). 
The goal of diffusion model inversion is to recover the initial noise \(z_T^*\) such that the reconstructed image \(x_0\) matches the input \(x_0\). This can be formulated as:  
\vspace{-0.5em}
\begin{equation}
z_T^* = \underset{z_T}{\mathrm{argmin}} \, \mathcal{L}(z_T, x_0),
\end{equation}  
or further decomposed into intermediate optimization problems at each timestep \(t\):  

\begin{equation}
z_t^* = \underset{z_t}{\mathrm{argmin}} \, \mathcal{L}(z_t, x_0).
\end{equation}  
Diffusion model inversion involves iteratively optimizing the intermediate latent variables \(z_t\) at each timestep, eventually recovering the initial noise \(z_T\). This process can also incorporate the denoising likelihood \(p_\theta(z_{t-1} | z_t)\) for timestep-by-timestep optimization. 
Detailed methods in \cref{fig:taxonomy_of_GMI} are detailed in the following sections

\subsection{GAN Inversion}
GAN inversion techniques can be broadly classified into three categories: 
\textit{i)} Encoder-based approaches prioritize speed and conditional control, \textit{ii)} Latent Optimization approaches emphasize reconstruction fidelity, and \textit{iii)} Hybrid approaches seek to balance these competing objectives. This categorization reflects the primary strategy employed to map a real image back into the latent space of a pre-trained GAN.

\nibf{Encoder-based Approaches.} 
In encoder-based image inversion methods, researchers focus on constructing a direct mapping from image space to the GAN latent space. These methods are renowned for their efficient inference speed and practical utility, making them particularly suitable for real-time processing scenarios and conditional generation tasks. A seminal work in this domain, IcGAN~\cite{IcGAN}, innovatively designed a dual-encoder architecture responsible for extracting the latent features $\mathbf{z}$ and the associated conditional information $\mathbf{y}$ from images. This breakthrough enabled conditional image editing based on cGANs~\cite{cGANs}. Subsequently, e4e~\cite{e4e} proposed a specialized encoding scheme tailored to the characteristics of StyleGAN~\cite{StyleGAN}. The core of this approach lies in the progressive offset training strategy combined with adversarial constraints, effectively bridging the gap between latent encoding and the native StyleGAN space, thereby significantly enhancing the model's editability and fidelity. Recent advancements such as DIEGAN~\cite{DIEGAN} employ a hybrid network that combines encoder-derived latent codes with randomly sampled codes for diverse image restoration and editing tasks, utilizing a two-stage training framework. Additionally, GradStyle~\cite{GradStyle} introduces a dual-stream framework integrating progressive residual alignment and global alignment modules, aimed at high-fidelity image reconstruction and flexible attribute manipulation. Overall, these technical approaches rely on establishing direct image-to-latent space mapping to achieve rapid inversion operations. In practical development, it remains crucial to carefully balance reconstruction quality and editability.

\nibf{Latent Optimization Approaches.} 
Latent optimization methods directly optimize the latent variables of a pre-trained GAN to find the optimal latent code reconstruct a given image, achieving high reconstruction fidelity without additional modules. InvertingGAN~\cite{InvertingGAN} uses gradient descent to directly invert GAN generators, achieving high-quality image reconstruction. Image2StyleGAN~\cite{Image2StyleGAN} embeds images into StyleGAN's extended W+ space via optimization, improving embedding quality and style transfer capabilities. PTI~\cite{PTI} combines latent space optimization with local generator fine-tuning, balancing reconstruction quality and editability by adapting the generator to specific input image characteristics.  These approaches prioritize reconstruction accuracy by iteratively refining latent codes, albeit at the cost of increased computational time. 
Considering the latency requirements of practical applications and the emergence of a series of high-quality diffusion generative models, such approaches have gradually faded from researchers' focus in recent years.

\nibf{Hybrid Approaches.} 
Hybrid methods combine the strengths of both encoder-based and latent optimization approaches, typically using an encoder to provide an initial estimate of the latent code, which is then refined through optimization, allowing a balance between speed and reconstruction quality, while also enabling more complex editing tasks. iGAN~\cite{iGAN} combines deep learning predictors and optimization techniques to efficiently project images onto the GAN manifold and enable interactive editing. SDIC~\cite{SDIC} learns a spatial-contextual discrepancy map to compensate for information loss in latent codes and GAN generators, generating high-quality reconstructions and edits. LOR~\cite{LOR} combines encoders and latent space direction optimization for high-quality face reenactment via joint training and feature space refinement. StyleFeatureEditor~\cite{StyleFeatureEditor} adopts a two-phase learning approach to train an inverter and a feature editor, addressing detail reconstruction and editability in high-dimensional feature spaces. These approaches leverage the efficiency of encoders and the precision of latent optimization, offering a versatile solution for GAN inversion and image editing.

\nibf{Pros and Cons Discussion.}
GAN inversion for image editing faces challenges in balancing reconstruction fidelity, editability, and computational efficiency. The field is moving towards hybrid methods that merge the speed of encoders with the precision of latent optimization. These approaches aim to enable real-time, high-quality image manipulation, preserving essential details and supporting complex editing tasks.

\subsection{Diffusion Model Inversion.}
According to the differences in training strategies, the Inversion of Diffusion Model for Image methods can be categorized into three types: training-free methods, fine-tuning methods, and methods with additional trainable modules. This classification provides a clear reflection of the differences in inversion strategies with respect to model training.

\nibf{Training-free Approaches.} 
Training-free methods aim to enable efficient image inversion and editing by directly utilizing pre-trained diffusion models, without relying on additional training or fine-tuning. This category is highly regarded for its convenience and flexibility, especially in scenarios demanding high efficiency. Prompt-to-Prompt (P2P)\cite{p2p} exemplifies this, achieving intuitive text-driven image editing by manipulating cross-attention maps, eliminating the need for additional model adjustments and providing a direct means of control for text-guided image manipulation. Building upon this, Null-Text Inversion\cite{Null-Text} leverages DDIM inversion, reconstructing and editing images by optimizing unconditional text embeddings, avoiding dependence on specific text information and enhancing editing flexibility. To achieve more precise inversion, EDICT~\cite{EDICT} introduces coupling transformations and mixing layers, aiming to improve the fidelity of the inversion results. Negative-prompt Inversion~\cite{Negative-prompt} takes a different approach, replacing null-text embeddings with negative-prompt embeddings, achieving efficient inversion and editing without optimization, simplifying the operation process. Furthermore, TIC~\cite{TIC} enhances features of DDIM inversion to improve editing consistency, ensuring the naturalness of the edited results. StyleID~\cite{StyleID} achieves training-free style transfer through self-attention feature replacement and initial noise adjustment, making style conversion more convenient. TF-ICON~\cite{TF-ICON} leverages exceptional prompts and attention injection to achieve cross-domain image composition, providing more possibilities for image creation. FreePromptEditing~\cite{FreePromptEditing} achieves efficient text-guided editing by modifying self-attention maps, improving editing efficiency. DesignEdit~\cite{DesignEdit} proposes a multi-layered latent decomposition and fusion framework, aiming to improve editing quality, making the editing results more refined. FreeControl~\cite{FreeControl} introduces spatial control, utilizing feature subspaces for spatial adjustment of text-to-image generation, providing stronger control capabilities for image generation. Blended Diffusion~\cite{Blended} blends text-guided diffusion latents with input noise, enabling seamless local image editing, making local editing more natural. InterpretDiffusion~\cite{InterpretDiffusion} optimizes latent vectors, exploring semantic directions applicable to fair, safe, and responsible generation, providing technical support for the ethical application of generative models. BIRD~\cite{BIRD} achieves training-free blind image restoration by optimizing the initial noise of pre-trained diffusion models, providing new ideas for image restoration. Therefore, training-free methods hold a significant position in image inversion due to their high efficiency and flexibility, along with continuously enriching technical means.

\nibf{Fine-tuning Approaches.} 
Fine-tuning methods improve the adaptability and performance of pre-trained diffusion models on specific tasks by adjusting their parameters. Unlike training-free methods, fine-tuning methods focus on targeted optimization of the model to adapt to specific application scenarios. DreamBooth~\cite{DreamBooth} is a representative example, binding specific subjects with rare identifiers by fine-tuning a small set of input images and introducing class-related prior preservation loss to enhance generation diversity, thus excelling in personalized generation. Forgedit~\cite{Forgedit} improves fine-tuning strategies with vector projection, achieving efficient text-guided image editing and enhancing editing efficiency. Photoswap~\cite{Photoswap}, building on this, combines attention swapping and fine-tuned subject concept learning for more personalized subject replacement. FaceChain-SuDe~\cite{FaceChain-SuDe} introduces Subject-Derived regularization (SuDe), enabling models to inherit common attributes of their category while retaining individual characteristics by fine-tuning partial parameters, thus possessing unique advantages in fields such as face editing. CLiC~\cite{CLiC} introduces innovative fine-tuning and cross-attention optimization, enabling in-context learning and transfer of local visual concepts, providing stronger semantic understanding capabilities for image editing. As such, fine-tuning methods can effectively enhance model performance on specific tasks, but also require a certain training cost, necessitating a trade-off between performance and cost.

\nibf{Extra Trainable Module.} 
Extra trainable module methods enhance the functionality of diffusion models by introducing new modules or adjusting network structures, particularly excelling in image editing tasks. Compared to fine-tuning methods, this category typically possesses greater scalability and flexibility, capable of handling more complex editing tasks. For example, StyleDiffusion~\cite{StyleDiffusion} optimizes input embeddings for cross-attention "value" layers, achieving text-guided localized or holistic style editing, providing new ideas for style transfer. DAC~\cite{DAC} introduces two LoRA modules within a doubly abductive reasoning framework to fine-tune the UNet and text encoder, thereby improving fidelity and editability, providing a guarantee for high-quality image editing. InST~\cite{InST} utilizes an attention-based textual inversion module for efficient artistic style transfer, further expanding the application scope of style transfer. DETEX~\cite{DETEX} achieves separation of background and pose information by fine-tuning subject embeddings on Stable Diffusion and introducing trainable attribute mappers, enabling flexible generation control, providing more refined control capabilities for image editing. TurboEdit~\cite{TurboEdit} trains encoder networks to reduce computational costs, thereby improving the efficiency of image editing. InvSR~\cite{InvSR} combines a noise predictor network with partial noise prediction strategies, achieving flexible and efficient super-resolution inversion, enhancing the resolution of image editing. Overall, extra trainable module methods provide diffusion models with greater scalability and flexibility, enabling them to better adapt to various image editing tasks, but also require more sophisticated design and optimization.

\nibf{Pros and Cons Discussion.}

Diffusion model inversion faces challenges in balancing reconstruction accuracy with editability, particularly in maintaining consistency and fidelity during manipulation. The field is progressing towards faster, more controllable, and inversion-free editing techniques, with a focus on improving the robustness and real-time applicability of these models for diverse image editing tasks.

\subsection{Promising Technologies on the Rise}

The latent space of GANs can be distorted, limiting editability, while Diffusion Models require many iterative steps, leading to high computational costs. To address these issues, Diffusion Transformers (DiT)~\cite{DiT} and Rectified Flows have emerged, offering new possibilities for image editing. 
\textbf{\textit{1) The DiT framework}}, with its global receptive field and strong modeling capabilities, excels in image generation and editing. Compared to GANs, DiT's latent space is smoother and more amenable to semantic operations, allowing for more controllable editing. Unlike traditional Diffusion Models, DiT achieves high-quality inversion with fewer steps, enhancing efficiency. For example, DiT4Edit~\cite{DiT4Edit} optimizes generation speed and editing quality by using DPM-Solver for fewer steps, unifying control of self- and cross-attention for layout and consistency, and introducing patch merging to reduce computation during inference. Stable Flow~\cite{Stable}, also based on DiT, selectively injects attention features into vital layers to maintain high-quality editing while preserving non-edited regions. 
\textbf{\textit{2) Rectified Flow framework}} avoids GAN latent space distortion and reduces Diffusion Model computational costs by directly mapping image space to latent space. RF Inversion~\cite{RFInversion} uses stochastic rectified differential equations and a dynamically controlled vector field for image inversion and editing without additional parameter training, balancing faithfulness and editability. GNRI~\cite{GNRI} improves text-to-image diffusion model inversion quality and speed by introducing a guidance regularization term into the Newton-Raphson optimization process. RF-Solver-Edit~\cite{RF-Solver-Edit}, using the training-free sampler RF-Solver, reduces ODE-solving errors with high-order Taylor expansion and achieves high-quality editing while preserving source image structure integrity. 
Overall, both above frameworks offer superior controllability and inversion efficiency, representing significant advancements and direction.

\section{Application Tasks} \label{sec:experiment}
\subsection{Popular Datasets}
\nibf{LAION}~\cite{LAION}
is a vast, open multimodal dataset with billions of image-text pairs across various languages, domains, and content. Its diversity is crucial for inversion tasks in generative models, particularly in cross-modal retrieval and generative AI training. Generative models can learn image-text relationships, facilitating the reverse mapping of semantic representations in latent space for text-guided image editing.

\nibf{PIE-bench}~\cite{TurboEdit}
is a specialized benchmark dataset for evaluating image editing models, comprising 700 images of natural and artificial scenes and covering 10 editing types like object replacement and style modification. Each entry includes detailed source and target image descriptions, editing instructions, and annotated editing masks. In generative model inversion, PIE-bench guides fine-grained latent space operations using these instructions and masks, allowing precise evaluation of image editing performance.

\nibf{FFHQ}~\cite{StyleGAN}
is a high-quality, diverse face dataset with 70,000 images up to 1024$\times$1024 resolution, encompassing various ages, genders, ethnicities, and backgrounds. It is extensively used in generative model inversion for facial generation and style transfer tasks. By inverting the latent vector, models can edit specific facial attributes (\textit{e.g.}, expression or hairstyle) while preserving other features.

\nibf{MS COCO}~\cite{MSCOCO}
offers 328,000 images across 91 categories in everyday scenes, with high-quality annotations for object detection, instance segmentation, panoptic segmentation, human keypoint detection, and image captioning. In generative model inversion, MS COCO's text descriptions and annotations assist models in locating specific object representations in the latent space, facilitating image editing and reconstruction in complex scenes.

\nibf{CelebA-HQ}~\cite{CelebA-HQ}
comprises 30,000 high-quality face images at 1024$\times$1024 resolution, annotated with 40 attribute labels (\textit{e.g.}, gender, age, hairstyle, and expressions). These labels enable precise control over attributes in generative model inversion. By editing the latent representation, models can manipulate specific attributes, such as changing hairstyles, while preserving other facial features.

\nibf{DreamBooth}~\cite{DreamBooth}
includes 3$\sim$5 high-quality images of 30 subjects and 25 text prompts for evaluating subject-driven generation tasks like recontextualization, property modification, accessorization, and artistic rendering. In generative model inversion, DreamBooth aids in exploring subject-specific representations in the latent space. By using unique identifiers and class names, the model enables personalized subject generation and attribute editing while preventing semantic drift. 

\subsection{Mainstream Applications}
Generative model inversion is pivotal in image editing and generation. By encoding real images into a latent space, image inversion methods leverage the capabilities of the latent space for manipulating images. We outlines the applications (summarized in \cref{table:application}) across various mainstream tasks without focusing on the implementation framework:

\begin{enumerate}[leftmargin=*,topsep=0pt,itemsep=0pt,align=left]
    \item \textbf{Object Editing:} Object editing enables the editing of specific objects within an image, including operations such as object replacement, removal, and addition. Methods include Prompt-to-Prompt~\cite{p2p} and Null-Text Inversion~\cite{Null-Text}.

    \item \textbf{Attribute Editing:} Attribute editing facilitates the modification of object attributes within an image, including editing the color, texture, and shape of objects, as well as the age, expression, and hairstyle of people.
        \begin{itemize}[leftmargin=*,topsep=0pt,itemsep=0pt,align=left]
            \item \textbf{Editing Direction:} Modifies attributes by adjusting the latent space vector of the image. Example methods: iGAN~\cite{iGAN}, Image2StyleGAN~\cite{Image2StyleGAN}, e4e~\cite{e4e}, \textit{etc}.
            \item \textbf{Prompt Editing:} Modifies attributes by modifying the text prompt. Example methods: HyperEditor~\cite{HyperEditor}, Prompt-to-Prompt~\cite{p2p}, Null-Text Inversion~\cite{Null-Text}.
        \end{itemize}

    \item \textbf{Style Transfer:} Style transfer allows for the transfer of style from one image to another, and it can also be achieved through Prompt-based style transfer, thereby changing the overall style of the image.
        \begin{itemize}[leftmargin=*,topsep=0pt,itemsep=0pt,align=left]
            \item \textbf{Based on Reference Image:} Changes the style of the target image using the style of a reference image. Example methods: Image2StyleGAN~\cite{Image2StyleGAN} and InST~\cite{InST}.
            \item \textbf{Based on Semantic Information:} Uses semantic information such as text prompts to control style transfer. Example methods: HyperEditor~\cite{HyperEditor}, Prompt-to-Prompt\cite{p2p}, Null-Text Inversion~\cite{Null-Text}.
        \end{itemize}

    \item \textbf{Image Concept Decoupling:} This task refers to learning local concepts from a single image and using them to generate other objects, thereby achieving independent control over different concepts in the image. CLiC~\cite{CLiC} is frequently employed.

    \item \textbf{Spatially Aware Editing:} Spatially aware editing enables editing images based on spatial information, including operations such as removal, movement, scaling, rotation, and translation. In addition to using Prompt control, Point drag control can also be used to achieve more refined spatial editing. DiffEditor~\cite{DiffEditor} and DesignEdit~\cite{DesignEdit} are commonly used methods.

    \item \textbf{Controllable Image Generation:} Allows users to control the image generation process through various inputs (\textit{e.g.}, sketches, segmentation maps, poses, etc.). Typical method is FreeControl~\cite{FreeControl}.

    \item \textbf{Personalized Generation:} Personalized generation can be used to reproduce specific subjects in different scenarios, such as generating images of specific subjects in different artistic styles, different perspectives, different accessories, different expressions, and different backgrounds.
        \begin{itemize}[leftmargin=*,topsep=0pt,itemsep=0pt,align=left]
            \item \textbf{Single-Sample Personalized Generation:} Personalizes generation using only one reference image. Example method: FaceChain-SuDe~\cite{FaceChain-SuDe}.
            \item \textbf{Multi-Sample Personalized Generation:} Personalizes generation using multiple reference images for better results. Typical methods: DreamBooth~\cite{DreamBooth} and DETEX~\cite{DETEX}.
        \end{itemize}

    \item \textbf{Image Restoration:} Image restoration can be used to repair damaged images, common sub-tasks are as follows:
    \begin{itemize}[leftmargin=*,topsep=0pt,itemsep=0pt,align=left]
        \item \textbf{Image Super-Resolution:} Increases the resolution of an image, making it clearer. Typical methods: BIRD~\cite{BIRD} and InvSR~\cite{InvSR}.
        \item \textbf{Denoising:} Removes noise from an image, making it cleaner. Typical method: BIRD~\cite{BIRD}.
        \item \textbf{Inpainting:} Repairs missing or damaged parts of an image. Typical methods: DIEGAN~\cite{DIEGAN}, and TurboEdit~\cite{TurboEdit}.
        \item \textbf{Deblurring:} Removes blur from an image, making it clearer. Typical method: BIRD~\cite{BIRD}.
    \end{itemize}

    \item \textbf{Image Fusion:} Image fusion refers to integrating content from two or more images into one image, thereby creating a new image. Photoswap~\cite{Photoswap}, TF-ICON~\cite{TF-ICON}, and TIGIC\cite{TIGIC} are commonly used methods.

\end{enumerate}

\begin{table}[tp]
    \centering
    \caption{Methods for downstream tasks. \gan{Rose red} and \dif{orange} colors represent GAN and diffusion inversion methods, respectively.}
    \label{table:application}
    \renewcommand{\arraystretch}{1.2}
    \setlength\tabcolsep{3.0pt}
    \resizebox{\columnwidth}{!}{ 
        \begin{tabular}{lp{0.9\linewidth}}
        \toprule[0.17em]
        \textbf{Task} & \textbf{Representative Methods} \\
        \midrule
        \makecell[l]{Object Editing} & \makecell[l]{\gan{iGAN}, \dif{Prompt-to-Prompt}, \dif{Null-Text Inversion} \\\dif{EDICT}, \dif{StyleDiffusion}, \dif{Negative-prompt Inversion} \\ \dif{Forgedit}, \dif{FreePromptEditing}, \dif{DAC}, \dif{TurboEdit}, \dif{TIC}} \\
        \midrule
        \makecell[l]{Attribute Editing} & \makecell[l]{\textbf{Editing Direction:} \gan{iGAN}, \gan{IcGAN}\\ \gan{Image2StyleGAN},\gan{ e4e},\gan{ PTI}, \gan{SDIC}, \gan{LOR}\\ \gan{GradStyle},\gan{ StyleFeatureEditor} \\ \textbf{Prompt Editing:} \gan{HyperEditor}, \dif{Prompt-to-Prompt}\\ \dif{Null-Text Inversion}, \dif{EDICT}, \dif{StyleDiffusion}\\ \dif{Negative-prompt Inversion}, \dif{Forgedit}\\ \dif{FreePromptEditing}, \dif{DAC}, \dif{TurboEdit}, \dif{TIC}} \\
        \midrule
        \makecell[l]{Style Transfer} & \makecell[l]{\textbf{Reference-Based:} \gan{Image2StyleGAN}, \dif{InST}, \dif{StyleID}\\ \textbf{Semantic-Based:} \gan{HyperEditor}, \dif{Prompt-to-Prompt}\\ \dif{Null-Text Inversion}, \dif{EDICT}, \dif{StyleDiffusion}\\ \dif{FreePromptEditing}, \dif{DAC}} \\
        \midrule
        \makecell[l]{Image Concept Decoupling} & \makecell[l]{\dif{CLiC}} \\
        \midrule
        \makecell[l]{Spatial-Aware Editing} & \makecell[l]{\dif{DiffEditor},\dif{ DesignEdit}} \\
        \midrule
        \makecell[l]{Controllable Image Generation} & \makecell[l]{\dif{FreeControl}} \\
        \midrule
        \makecell[l]{Personalized Generation} & \makecell[l]{\textbf{Single-Sample Personalization:} \dif{FaceChain-SuDe} \\ \textbf{Multi-Sample Personalization:} \dif{DreamBooth}, \dif{DETEX}} \\
        \midrule
        \makecell[l]{Image Restoration} & \makecell[l]{\textbf{Super-Resolution:} \dif{BIRD}, \dif{InvSR} \\ \textbf{Denoising:} \dif{BIRD} \\ \textbf{Inpainting:} \gan{DIEGAN}, \dif{TurboEdit}, \dif{Blended Diffusion} \\ \textbf{Deblurring:} \dif{BIRD}} \\
        \midrule
        \makecell[l]{Image Fusion} & \makecell[l]{\textbf{Subject Replacement:} \dif{Photoswap} \\ \textbf{Pasting:} \dif{TF-ICON}, \dif{TIGIC}} \\
        \bottomrule[0.17em]
        \end{tabular}
    }
\end{table}

\subsection{Related Domain Research}
To further enrich the research on inversion tasks, we briefly review related popular tasks beyond image inversion to strengthen this inspection. 

\nibf{Video Inversion.}
In the video domain, RIGID~\cite{RIGID} introduces an additional trainable temporal encoder combined with self-supervised constraints to learn inter-frame temporal coherence, enabling high-quality video reconstruction and editing without retraining the generator. Built on StyleGAN2~\cite{StyleGAN2}, it optimizes frame consistency and global identity preservation. Meanwhile, Videoshop~\cite{Videoshop} allows users to edit the first frame of a video and automatically propagates the changes across the entire sequence using noise extrapolation and latent normalization techniques, maintaining semantic consistency, motion naturalness, and geometric stability.

\nibf{Audio Inversion.}
MusicMagus\cite{MusicMagus} is a zero-shot music editing method that leverages pretrained diffusion models with latent space manipulation and cross-attention constraints, enabling text-guided audio editing without additional training. Similarly, ZETA~\cite{ZETA} introduces a zero-shot audio editing method using diffusion model inversion to extract noise semantic directions for fine-grained audio editing, supporting tasks such as text-guided removal, style changes, improvisation, and pitch adjustments.

\nibf{3D Inversion.}
In the 3D domain, E3DGE~\cite{E3DGE} is an encoder-based method that leverages self-supervised learning with global latent codes and pixel-aligned local features for high-fidelity 3D face reconstruction and consistent editing. Its dual-module structure includes global and local encoders to learn coarse 3D shapes and detailed textures, ensuring high consistency during viewpoint changes and semantic editing. Meanwhile, SHAP-EDITOR\cite{SHAP-EDITOR} trains a universal latent editor network by distilling knowledge from 2D diffusion models (\textit{e.g.}, InstructPix2Pix~\cite{InstructPix2Pix}), transferring the understanding of natural language instructions into the 3D latent space for fast and high-quality 3D editing.

\nibf{Medical Image.}
In the medical domain, TPDM\cite{TPDM} leverages two perpendicular 2D diffusion models to jointly model 3D medical data distributions, achieving significant improvements in tasks such as Z-axis super-resolution, compressed sensing MRI, and sparse-view CT reconstruction. InverseSR~\cite{InverseSR}, on the other hand, optimizes latent codes to adapt to both high-sparsity and low-sparsity scenarios, enabling high-resolution MRI reconstruction from low-resolution MRI, excelling in tasks like slice imputation and tumor/lesion filling.

\section{Open Problems and Promising Directions} \label{sec:experiment}
\subsection{Open Problems}
\nibf{Trading-off high-fidelity reconstruction and controllable editing.} 
GAN/Diffusion inversion often suffers from ``overfitting" the latent space when pursuing pixel-level high-fidelity reconstruction, which sacrifices the ability for subsequent controllable editing, and vice versa. There is a need to define more reasonable latent space constraints while exploring more expressive and generalized large-scale generative models trained on large datasets. 

\nibf{Computational efficiency bottleneck.} 
Although GAN inversion via encoders is faster, it often sacrifices precision. In contrast, diffusion model inversion achieves better results but is computationally expensive due to iterative optimization, especially for high-resolution generation. This increases training/inference costs and limits applications in resource-constrained scenarios.

\nibf{Challenges in multimodal and cross-domain inversion}
Existing inversion methods are primarily focused on single modalities and struggle to generalize to multimodal (\textit{e.g.}, image-to-text) or cross-domain tasks (\textit{e.g.}, medical image to natural image). Developing universal inversion frameworks capable of handling multimodal inputs while maintaining semantic consistency is critical.

\nibf{Data bias and fairness issues}
Generative model inversion may produce biased or unfair results due to data imbalances, such as under- or over-representing specific racial or gender attributes. Introducing fairness constraints into the inversion process to mitigate bias is an urgent problem to address.

\subsection{Promising Directions}

\nibf{Multi-scale latent space representation learning.} 
By introducing multi-scale latent space representations (\textit{e.g.}, combining local and global encodings), a better balance between high-fidelity reconstruction and controllable editing can be achieved. Further research on defining semantically consistent latent space representations will be crucial.

\nibf{Efficient optimization and inference algorithms.}
To address the iterative optimization bottleneck, lightweight optimization modules or efficient algorithms based on approximate inference (\textit{e.g.}, explicit latent space mapping or hierarchical optimization methods) can significantly reduce computational costs and improve efficiency in high-resolution scenarios.

\nibf{Multimodal consistency inversion techniques.}
Developing inversion frameworks that support multimodal inputs, such as jointly optimizing the latent spaces of images and texts, can enhance performance in cross-modal tasks. Ensuring semantic consistency in multimodal generation tasks (\textit{e.g.}, image-to-text-to-image) will be a key focus.

\nibf{Autoregressive image generation inversion framework.}
Recent autoregressive-based image and video generation has gained attention. Compared to diffusion models, autoregressive models offer higher efficiency and are expected to become one of next-generation foundational models.

\nibf{Inversion research based on Chain-of-Thought (CoT).}
CoT technology significantly enhances the reasoning capabilities of large language models. The incorporation of reinforcement learning enables the generation of results that better align with human thought processes. Recently, this technology has also been extended to the field of image generation. Inversion schemes based on CoT can further improve the quality and controllability of generated images.

\section{Conclusion} \label{sec:conclusion}

This survey comprehensively reviews the latest advancements in the field of image inversion, summarizing the inversion methods of GANs and diffusion models, along with their key applications. Specifically, this paper delves into the fundamental theoretical frameworks, optimization methods, and the evolution of various techniques for GAN and diffusion model inversion. It also introduces the extensive applications of these methods in tasks such as image editing, style transfer, and image restoration. Additionally, we discuss related fields beyond images, summarize the current challenges faced by inversion techniques, and propose directions for future research.

\bibliographystyle{named}
\bibliography{IEEEabrv,ijcai25}

\end{document}